\pdfoutput=1

\documentclass[11pt]{article}

\usepackage[preprint]{acl}

\usepackage{times}
\usepackage{latexsym}

\usepackage[T1]{fontenc}

\usepackage[utf8]{inputenc}

\usepackage{microtype}

\usepackage{inconsolata}

\usepackage{graphicx}
\usepackage{amssymb}
\usepackage{booktabs}
\usepackage{multicol}
\usepackage{multirow}

\usepackage{url}

\usepackage{pgfplots,pgfplotstable}
\tikzset{
  font={\fontsize{8pt}{10}\selectfont}}
\usepackage{graphicx}
\usepackage{subfigure}
\usepackage{tikz}
\usetikzlibrary{patterns, patterns.meta}
\pgfplotsset{compat=1.18}

\pgfplotsset{
  /pgfplots/colormap={rwg}{
    rgb255(0pt)=(130,20,20)
    rgb255(50pt)=(220,60,60)
    rgb255(125pt)=(255,255,255)
    rgb255(200pt)=(60,180,90)
    rgb255(250pt)=(20,110,45)
  }
}

\definecolor{darkyellow}{RGB}{251,188,4}
\definecolor{darkgreen}{RGB}{52,168,83}
\definecolor{lightblue}{RGB}{66,133,244}
\definecolor{acqua}{RGB}{70,189,198}

\usepackage[most]{tcolorbox}
\usepackage{adjustbox}
\tcbset{
  colframe=gray!30,
  colback=gray!4,
  boxrule=0.4pt,
  arc=2pt,
  outer arc=1pt,
  left=4pt,
  right=4pt,
  top=4pt,
  bottom=4pt,
  fonttitle=\bfseries,
  coltitle=black,
  enhanced jigsaw,
  breakable,
}
\newtcolorbox{promptbox}[2][]{
  title={#2},
  coltitle=black,
  fontupper=\ttfamily\small,
  #1
}

\title{DOA: Training-Free Decoder-Only Attention Policy for Long-Form Simultaneous Translation with SpeechLLMs}

\author{Sara Papi \and Luisa Bentivogli\\
  Fondazione Bruno Kessler, Italy \\
  \texttt{\{spapi,bentivo\}@fbk.eu}}

\begin{document}
\maketitle
\begin{abstract}
Simultaneous speech-to-text translation (SimulST) generates translations while speech is still unfolding, requiring a streaming policy that decides when to read and when to write. State-of-the-art approaches rely on attention-based encoder-decoder models where cross-attention provides explicit alignment signals. In contrast, Speech Large Language Models (SpeechLLMs) are decoder-only architectures relying solely on self-attention. This raises a central question: whether decoder self-attention contains sufficiently stable alignment signals to guide the streaming policy. Moreover, existing approaches typically rely on training-based adaptations or heuristic wait-$k$ policies and have not been validated in long-form settings. To fill these gaps, we propose \textbf{D}ecoder-\textbf{O}nly \textbf{A}ttention (DOA), a training-free policy that enables long-form simultaneous translation with off-the-shelf SpeechLLMs by deriving a proxy alignment from self-attention. Experiments on Phi4-Multimodal and Qwen3-Omni show that DOA provides an effective alignment signal for supporting streaming decisions, enabling low-latency long-form SimulST with quality close to offline decoding without retraining.
\end{abstract}

\section{Introduction}

Simultaneous speech-to-text translation (SimulST) aims to generate translations while the source speech is still unfolding, balancing translation quality and latency \citep{fugen2007simultaneous}. This requires a \emph{simultaneous policy}, i.e., a decision strategy that determines when to read additional speech input and when to write output tokens during streaming inference \citep{grissom-ii-etal-2014-dont}.  

Recent advances in Speech Large Language Models (SpeechLLMs) have shifted attention toward decoder-only architectures, which unify speech and text processing within a single autoregressive LLM-powered model \citep{wu2023decoder,huang24h_interspeech}. While these models have shown strong performance across speech understanding and offline translation tasks \citep{huang24h_interspeech,gupta24_interspeech,papi2026hearingtranslate}, their applicability to SimulST remains largely unexplored. Existing work on SpeechLLMs either relies on ad-hoc training or fine-tuning to induce streaming behavior \citep{10832146,guo2025streamuniachievingstreamingspeech,ouyang2024fasst,ouyang-etal-2025-infinisst}, or adapts classical training-free heuristics such as wait-$k$ policies \citep{ma-etal-2019-stacl} to decoder-only architectures \citep{Guo_Zhang_Ma_Feng_2025}.
In contrast, state-of-the-art systems in encoder-decoder settings are typically driven by training-free attention-based policies that exploit cross-attention signals to drive streaming decisions, rather than relying on fixed schedules \citep{ahmad-etal-2024-findings,agostinelli-etal-2025-findings}.

At the same time, previous studies have shown that evaluating SimulST systems on pre-segmented utterances can underestimate the challenges of realistic streaming conditions \citep{papi-etal-2025-real}, where models must continuously process growing acoustic and textual contexts over long-form audio streams \citep{polak2023long}. Despite these challenges, no prior work has investigated whether offline SpeechLLMs can be directly repurposed for long-form SimulST through a training-free attention-based policy.

A key obstacle is that SpeechLLMs do not expose explicit cross-attention scores. In attention-based encoder-decoder (AED) systems, cross-attention provides alignment signals that are sufficiently stable to support streaming decisions \citep{papi-etal-2023-attention,papi23_interspeech,wang24ea_interspeech}. However, it remains unclear whether SpeechLLMs' self-attention exhibits similar properties. This motivates the main research question of this work: \emph{does SpeechLLMs' self-attention provide sufficiently stable alignment information to support streaming decisions as in AED models?}



To answer this question, we propose \textbf{D}ecoder-\textbf{O}nly \textbf{A}ttention (DOA), the first training-free attention-based policy for long-form simultaneous translation with decoder-only SpeechLLMs.\footnote{Code is released at \url{https://github.com/hlt-mt/simulstream} under Apache 2.0 License.} DOA derives a proxy cross-attention matrix directly from decoder self-attention weights and exploits the resulting alignments to drive incremental generation while dynamically pruning both acoustic and textual histories during streaming inference for efficient long-form processing. 

Experiments on popular off-the-shelf SpeechLLMs, Phi4-Multimodal and Qwen3-Omni, on English$\rightarrow$German and English$\rightarrow$Italian translation show that DOA can effectively generalize beyond encoder-decoder architectures, achieving low-latency streaming generation with quality close to offline decoding without task-specific retraining.

\section{Decoder-Only Attention Policy}

\subsection{Simultaneous Policy}

We propose a \textbf{d}ecoder-\textbf{o}nly \textbf{a}ttention-based (DOA) streaming policy for SpeechLLMs, adapting the policy originally introduced for attention-based encoder-decoder (AED) models \citep{papi-etal-2024-streamatt} to architectures without explicit cross-attention. Unlike AED systems, decoder-only models process speech and text within a single autoregressive sequence, making source-target alignments less explicit.
To address this limitation, we derive a \emph{proxy cross-attention} signal from decoder self-attention weights. During autoregressive generation, we extract decoder self-attention weights:
\[
A^{(l,h)} \in \mathbb{R}^{T \times (S+T)},
\]
where $l$ is the decoder layer, $h$ is the attention head, and $S$ and $T$ denote the number of audio and generated text tokens, respectively. Since audio tokens occupy the prefix of the sequence, we isolate attention directed toward the acoustic context:
\[
\tilde{A}^{(l,h)} = A^{(l,h)}[:, :S],
\]
obtaining a proxy cross-attention matrix
\[
\tilde{A} \in \mathbb{R}^{T \times S},
\]
analogous to the cross-attention matrix in AED models.

The attention matrix can be extracted from a single layer/head or averaged across selected layers and heads:
\[
\bar{A} =
\frac{1}{|\mathcal{L}||\mathcal{H}|}
\sum_{l \in \mathcal{L}}
\sum_{h \in \mathcal{H}}
\tilde{A}^{(l,h)}.
\]

Inspired by prior works \citep{papi23_interspeech,wang24ea_interspeech}, the resulting matrix is converted into a token-to-audio alignment by selecting, for each generated token, the audio position receiving the highest attention score:
\[
a_t = \arg\max_s \bar{A}_{t,s},
\]
where $a_t$ denotes the aligned audio index for token $t$. The alignment sequence is then used by the streaming policy to estimate whether newly generated tokens are grounded in the currently available audio context.
We introduce a cutoff hyperparameter $f$ representing the number of most recently received audio frames considered acoustically unstable. Tokens aligned to positions falling within the last $f$ frames are not emitted, as their supporting acoustic evidence may still evolve with future input. Therefore, only tokens satisfying\[a_t < S - f\] are committed to the output, where $S$ denotes the current number of audio frames. Larger values of $f$ increase the amount of audio treated as unstable, leading to more conservative generation and higher latency, while smaller values favor lower latency at the risk of emitting less stable hypotheses.

\subsection{Long-form Adaptation}

The DOA policy operates incrementally over a rolling, potentially extremely long, audio sequence. At each step, the newly received speech chunk is appended to the previously observed waveform, the \textit{audio history}, and provided to the model together with a retained \textit{textual history} used as prefix for the next decoding step. To avoid both histories to grow indefinitely, audio and textual history selection mechanisms are adopted.

Following prior streaming frameworks \citep{iranzo-sanchez-etal-2022-simultaneous,10.1162/tacl_a_00691,papi-etal-2024-streamatt}, we explore two history selection strategies for the textual part: \textit{(i)} \textbf{fixed words}, which preserves the last $N$ generated words/characters, and \textit{(ii)} \textbf{punctuation}, which preserves only the text segment after the most recent strong punctuation mark.

The proxy cross-attention-based alignments are then used for the audio history selection. Specifically, we discard consecutive audio frames aligned exclusively with the discarded textual history, while retaining the frames aligned with the preserved textual prefix and the newly generated hypothesis. This mechanism progressively removes acoustic segments already translated and no longer required by the model, enabling long-form streaming inference without unbounded context growth.
As an edge case, where attention-based pruning fails to discard any audio frames and the audio history exceeds a predefined duration, the oldest frames are truncated to maintain bounded memory usage.

Our framework is completely \textit{model-agnostic} and only requires access to decoder self-attention weights, making it applicable to any SpeechLLM.

\section{Experimental Settings}
\paragraph{Data and Metrics.} Following standard evaluation settings of the IWSLT Evaluation Campaigns \citep{agostinelli-etal-2025-findings},  we adopt MCIF \citep{papi2026mcif} as test set for English$\rightarrow$German and English$\rightarrow$Italian, and ACL 60/60 \citep{salesky-etal-2023-evaluating} English$\rightarrow$German as dev set for the hyperparameters selection and analyses. Always following IWSLT 2026 settings, we use the SimulStream toolkit \citep{gaido2025simulstream} as the inference framework, and OmniST-Eval \citep{polak2025better} 
to compute the LongYAAL and LongLAAL latency metrics. BLEU and COMET \citep{rei-etal-2022-comet} are used as quality metrics.

\paragraph{Models.} For the experiments, we selected open-weight SpeechLLMs based on their language support, therefore capable of translating from English into German and Italian.  This resulted into the selection of Phi4-Multimodal \citep{microsoft2025phi4} and Qwen3-Omni \citep{xu2025qwen3omni}. Hyperparameters selection and analyses are performed on Phi4-Multimodal only, and the model-agnosticity is verified on Qwen3-Omni in the final results. We also compare with the SimulStream baseline, StreamAtt \citep{papi-etal-2024-streamatt}, applied to the AED SeamlessM4T model \citep{seamlessm4t}. Detailed settings 
are in Appendix \ref{app:exp-sett}.

\section{Results}

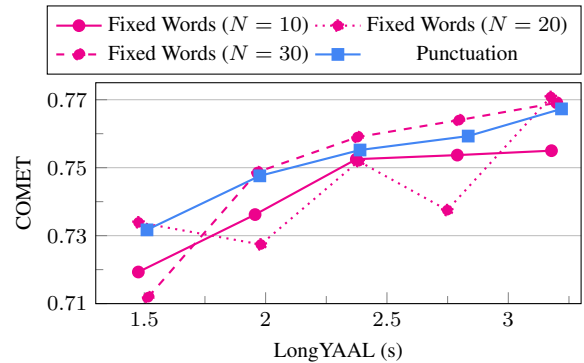
\begin{figure}[!ht]
\centering
\small
\begin{tikzpicture}
    \begin{axis}[
            ymajorgrids=true,
            xtick pos=left,
            ytick pos=left,
            minor y tick num=1,
            minor x tick num=1,
            ytick={0.71,0.73,0.75,0.77},
            ymin=0.71,
            ymax=0.775,
            xmin=1.3,
            xmax=3.3,
            ylabel=COMET, xlabel=LongYAAL (s),
            width=8cm,
            height=4.5cm,
            xtick=data,
            compat=newest,
            xtick={1.5,2,2.5,3,3.5,4},
            every axis plot/.append style={thick},
            legend style={at={(0.45,1.35)},    
                    anchor=north,legend columns=2},
        ]
        \addplot[color=magenta, mark=*] table[x=LAAL,y=COMET]{\fixedwords};
        \addplot[color=magenta, mark=*, dotted] table[x=LAAL,y=COMET]{\fixedwordstwenty};
        \addplot[color=magenta, mark=*, dashed] table[x=LAAL,y=COMET]{\fixedwordsthirty};
        \addplot[color=lightblue, mark=square*] table[x=LAAL,y=COMET]{\punctuation};
        \legend{Fixed Words ($N=10$),Fixed Words ($N=20$),Fixed Words ($N=30$),Punctuation}
    \end{axis}
\end{tikzpicture}
\caption{Latency (LongLAAL$\downarrow$) - Quality (COMET$\uparrow$) curves of Punctuation and Fixed Words methods applied to Phi4-Multimodal on ACL 60/60 en-de dev set. Numerical results are in Appendix \ref{app:final-table}.}
\label{fig:punct-vs-fix}
\end{figure}

\paragraph{Punctuation vs. Fixed Textual History Selection.} 
Figure \ref{fig:punct-vs-fix} shows the results of the DOA policy applied to Phi4-Multimodal with the two textual history selection methods, Fixed Words and Punctuation, with proxy cross-attention matrices obtained by averaging across layers and heads. The curves clearly show that the Punctuation method is more stable, yielding the best or near-best quality across all latency regimes, while remaining comparable in terms of latency (spanning between 1.5 and 3.5$s$). Interestingly, this behavior contrasts with current findings on AED-based streaming policies, where Fixed Words generally outperformed Punctuation-based history selection \citep{papi-etal-2024-streamatt}. 
Prior work has shown that punctuation and sentence segmentation are important for improving machine translation quality \citep{cho17_interspeech,cho-etal-2017-domain}, and that contextual continuity is beneficial in transformer-based speech sequence modeling \citep{10.1162/tacl_a_00420,huang23b_interspeech}. Consistent with these findings, our results suggest that decoder-only SpeechLLMs benefit from punctuation-based history selection, which preserves sentence-level textual continuity and yields a more stable autoregressive decoding context than retaining a fixed number of words.
These results motivate the adoption of the Punctuation strategy throughout the remainder of the paper.

\pgfplotstableread[row sep=\\]{
COMET	LAAL \\
0.7317	1.5111261\\
0.7476	1.9760655\\
0.7552	2.3889781\\
0.7593	2.8348213\\
0.7673	3.219304\\
}\punctuation

\pgfplotstableread[row sep=\\]{
COMET	LAAL \\
0.7193	1.4755524\\
0.7362	1.9565842\\
0.7525	2.373716\\
0.7537	2.7898177\\
0.755	3.1779547\\
}\fixedwords

\pgfplotstableread[row sep=\\]{
COMET	LAAL \\
0.7339	1.4762357\\
0.7274	1.9801015\\
0.7519	2.3834853\\
0.7375	2.7484396\\
0.7708	3.1752955\\
}\fixedwordstwenty

\pgfplotstableread[row sep=\\]{
COMET	LAAL \\
0.7119	1.516037\\
0.7487	1.9660534\\
0.759	2.3786357\\
0.764	2.7965606\\
0.7691	3.1997589\\
}\fixedwordsthirty

\begin{figure*}[!h]
\subfigure[Layer-wise Performance]{
    \centering
    \includegraphics[width=0.95\linewidth]{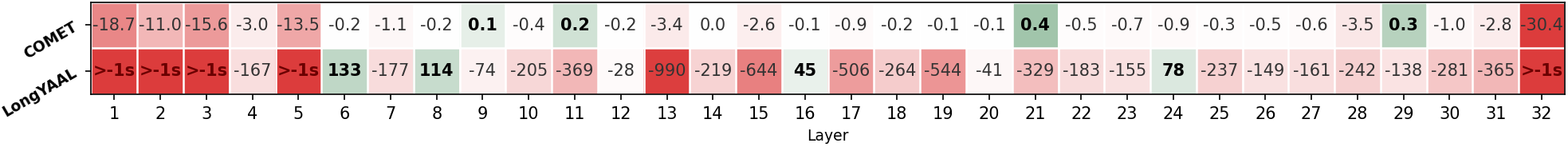}
    }
\subfigure[Head-wise Performance]{
    \centering
    \hspace{1.5cm}
    \includegraphics[width=0.8\linewidth]{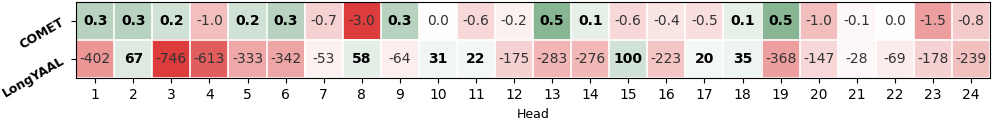}
    \hspace{1.5cm}
}
    \caption{Layer- and Head-wise performance difference compared to the average. Green squares indicate improvement, and red degradation, with their magnitude (COMET $\times$100 and LongYAAL in $ms$ for readability).}
    \label{fig:layer-head}

\end{figure*}

\pgfplotstableread[row sep=\\]{
COMET	LAAL \\
0.7091     1.3382115 \\
0.7602	2.2404688 \\
0.7682	3.0444913 \\
}\DEphi

\pgfplotstableread[row sep=\\]{
COMET	LAAL \\
0.7392	0.724852 \\
0.7884	2.7438001 \\
0.7911 3.7489038 \\ 
}\DEqwen

\pgfplotstableread[row sep=\\]{
COMET	LAAL \\
0.6857	1.6821241 \\
0.7026	2.2654729 \\
0.7124	3.1574615 \\
}\DEsmt

\pgfplotstableread[row sep=\\]{
COMET	LAAL \\
0.7707	1.3268348 \\
0.787	2.173599 \\
0.7985	3.0240741 \\
}\ITphi

\pgfplotstableread[row sep=\\]{
COMET	LAAL \\
0.801	0.0718073 \\
0.8086	0.6201064 \\
0.8282	3.8055494 \\
}\ITqwen

\pgfplotstableread[row sep=\\]{
COMET	LAAL \\
0.7592	1.6313071 \\
0.7663	2.2381292 \\
0.7725	3.0644588 \\
}\ITsmt

\begin{figure*}[!ht]
\centering
\small
\subfigure[en$\rightarrow$de]{
\begin{tikzpicture}
    \begin{axis}[
            ymajorgrids=true,
            xtick pos=left,
            ytick pos=left,
            minor y tick num=1,
            minor x tick num=1,
            ymin=0.68,
            ymax=0.8,
            xmin=0,
            xmax=4,
            ylabel=COMET, xlabel=LongYAAL (s),
            width=7.25cm,
            height=4.5cm,
            xtick=data,
            compat=newest,
            xtick={0,1,2,3,4},
            ytick={0.68,0.7,0.72,0.74,0.76,0.78,0.8},
            every axis plot/.append style={thick}
        ]
        \addplot[color=magenta, mark=square*] table[x=LAAL,y=COMET]{\DEqwen};
        \addplot[color=lightblue, mark=*] table[x=LAAL,y=COMET]{\DEphi};
        \addplot[color=darkgreen, mark=triangle*] table[x=LAAL,y=COMET]{\DEsmt};
    \end{axis}
\end{tikzpicture}
}
\qquad
\subfigure[en$\rightarrow$it]{
\begin{tikzpicture}
    \footnotesize
    \begin{axis}[
            ymajorgrids=true,
            xtick pos=left,
            ytick pos=left,
            minor y tick num=1,
            minor x tick num=1,
            ymax=0.84,
            ymin=0.75,
            xmin=0,
            xmax=4,
            ylabel=COMET, xlabel=LongYAAL (s),
            width=7.25cm,
            height=4.5cm,
            compat=newest,
            xtick=data,
            xtick={0,1,2,3,4},
            ytick={0.74,0.76,0.78,0.8,0.82,0.84},
            legend style={at={(0.5,-0.2)},    
                    anchor=north,legend columns=4},   
            legend to name={mylegend},
            every axis plot/.append style={thick}
        ]
        \addplot[color=lightblue, mark=*] table[x=LAAL,y=COMET]{\ITphi};
        \addplot[color=magenta, mark=square*] table[x=LAAL,y=COMET]{\ITqwen};
        \addplot[color=darkgreen, mark=triangle*] table[x=LAAL,y=COMET]{\ITsmt};
        \legend{Phi4-Multimodal, Qwen3-Omni, SeamlessM4T}
    \end{axis}
\end{tikzpicture} 
}
\ref{mylegend}
\caption{Latency (LongYAAL$\downarrow$) - Quality (COMET$\uparrow$) curves on MCIF of DOA policy on Phi4-Multimodal and Qwen3-Omni, and of StreamAtt baseline on SeamlessM4T. Numerical results are in Appendix \ref{app:final-table}.}
\label{fig:final}
\end{figure*}
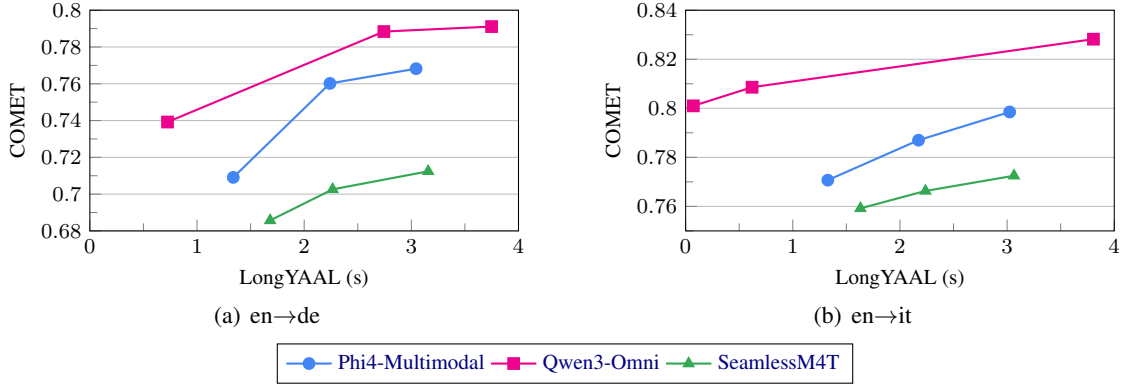

\paragraph{Layers and Heads Analysis.} Figure \ref{fig:layer-head} shows the performance difference between averaging the proxy cross-attention matrix across both layers and heads, and selecting a specific layer or head. Notably, the performance across layers is more unstable than across heads, with six layers leading to a complete failure (latency degradation close or more than 1$s$). No layer leads to gains on both latency and quality with respect to averaging, in contrast with previous works that found specific layer selection outperforms the average \citep{papi-etal-2023-attention}. Conversely, some heads improves in both quality and latency (e.g., heads 2 and 10) but gains are minimal. Therefore, the average across layers and heads represents the easiest
and best-performing choice, and it is used for final results.


\paragraph{Final Results.} Figure \ref{fig:final} reports the performance of the proposed DOA policy applied to both off-the-shelf Phi4-Multimodal and Qwen3-Omni using attention averaged across layers and heads. 
Both models achieve highly competitive translation quality, approaching the performance of offline systems reported on the same benchmark \citep{papi2026mcif}, where the best SpeechLLM (Phi4-Multimodal) reaches COMET scores of 0.78 on English-German and 0.81 on English-Italian. 
In terms of latency, Qwen3-Omni covers a broad operating range, spanning from 400$ms$ to 3.8$s$ average LongYAAL across languages, while also achieving the best overall translation quality. Phi4-Multimodal, although at lower COMET scores (with a difference of 0.02 on average) compared to Qwen3-Omni, is easier to manage in terms of latency, as increasing the cut-off frame by 10 always corresponds to 800$ms$ latency increase.
Notably, both DOA-equipped SpeechLLMs consistently outperform the StreamAtt AED baseline in terms of latency-quality trade-off, demonstrating the effectiveness of leveraging decoder self-attention for streaming decisions.
Finally, the strong performance obtained by both Phi- and Qwen-based models highlights the robustness and model-agnostic nature of DOA. Despite the substantial architectural differences between the two families (dense vs. MoE), the proposed policy generalizes effectively without requiring any adaptation.

\section{Conclusions}

We presented DOA, a training-free policy that enables long-form simultaneous speech-to-text translation with off-the-shelf decoder-only SpeechLLMs by exploiting self-attention as a proxy alignment signal. Our results show that \textbf{decoder self-attention provides sufficiently stable information to drive effective streaming decisions}, making it possible to repurpose offline SpeechLLMs for SimulST without retraining. We also find that punctuation-based history selection is consistently more effective than fixed-word strategies, and that averaging across layers and heads leads to the best results. These findings suggest that attention-based alignment can generalize beyond encoder-decoder models and serve as a simple yet effective mechanism for streaming inference in SpeechLLMs.

\section*{Limitations}
Our evaluation is limited to English source speech, primarily due to the scarcity of publicly available speech translation benchmarks with continuous audio of several minutes of duration. While this setting is representative of most long-form SimulST works \citep{papi-etal-2024-streamatt,ouyang-etal-2025-infinisst}, it does not fully capture multilingual or cross-domain variability in real-world deployments.

In addition, we focus on output languages using Latin scripts (German, Italian). It remains an open question whether the proposed proxy alignment derived from decoder self-attention generalizes equally well to languages with different scripts or tokenization characteristics, such as logographic (e.g., Chinese or Japanese) or morphologically rich languages (e.g., Turkish or Finnish).

Finally, we do not report computationally aware latency due to a heterogeneous GPU execution environment across experiments. While we provide \textit{ideal} latency comparisons by following standard best practices \citep{papi-etal-2025-real}, computational overhead and hardware requirements is substantially different across models (especially of Qwen3-Omni versus Phi4-Multimodal and SeamlessM4T), and therefore absolute timing comparisons across architectures should be interpreted with caution. Besides hardware inhomogeneity, other factors such as codebase optimization and different HuggingFace repository versions required by each model (see Appendix \ref{app:exp-sett}) can lead to significant computationally aware latency differences \citep{chitty2024llm}, and we believe that comprehensively analyzing these aspects is out of scope for this work.

\bibliography{custom}

\appendix

\clearpage

\begin{table*}[!ht]
\footnotesize
    \centering
    \setlength{\tabcolsep}{5pt}
    \begin{tabular}{llllllllllll}
    \toprule
    \textbf{Model} & \textbf{Param.} & \textbf{Weights} & \textbf{HFv} \\
    \midrule
    Phi4-Multimodal \citep{microsoft2025phi4} & 5.6B &   \href{https://hf.co/microsoft/Phi-4-multimodal-instruct}{\texttt{microsoft/Phi-4-multimodal-instruct}} & 4.48.2 \\
    Qwen3-Omni \citep{xu2025qwen3omni} & 30B & \href{https://huggingface.co/Qwen/Qwen3-Omni-30B-A3B-Instruct}{\texttt{Qwen/Qwen3-Omni-30B-A3B-Instruct}} & 5.0.0 \\
    SeamlessM4T \citep{seamlessm4t} & 1B &  \href{https://huggingface.co/facebook/hf-seamless-m4t-medium}{\texttt{facebook/hf-seamless-m4t-medium}} & 4.48.2 \\
    \bottomrule
    \end{tabular}
    \caption{Details of the analyzed models, including the number of parameters, their public weights release, and the HuggingFace Transformer version (HFv) used for the experiments.}
    \label{tab:models-details}
\end{table*}

\section{Detailed Experimental Settings}
\label{app:exp-sett}

Information about model version and weights are provided in Table \ref{tab:models-details}.

For SpeechLLMs, the incremental speech chunk size received at each step is set to 1 second, the default of SimulStream. To obtain quality-latency curves, typical of SimulST evaluation \citep{ma-etal-2020-simulmt}, we vary the cutoff frame $f$ in $\{5, 10, 15, 20, 25\}$ for Phi4-Multimodal during parameter selection.
For the final results, both Phi4-Multimodal and Qwen3-Omni cut-off frame is varied in $\{5, 15, 25\}$ for both target languages to obtain the three latency regimes: low, medium, and high latency. For the AED baseline, the SeamlessM4T v1 medium model, default settings from SimulStream are adoped and the cut-off frame is varied in $\{4, 8, 12\}$. 

The maximum audio length is set to 120$s$ for Phi4-Multimodal and SeamlessM4T, and 90$s$ Qwen3-Omni as a fallback in the edge cases in which attention does not discard enough frames and the history risks to exceed the available memory.
The maximum number of new tokens allowed to be generated at each step is set to $32$, and the maximum textual history length (in tokens) is $128$. 

Following the specific model cards, the prompts used for inference are the following(a simpler one for Phi4-Multimodal,\footnote{Preliminary experiments on Phi4-Multimodal with more complex prompts led to significant degradation in the results.} a more complex one for Qwen3-Omni):
\begin{promptbox}[colback=darkgreen!20!white, colframe=darkgreen!80]{Phi4-Multimodal}
Translate the audio to \textbf{\{tgt\_lang\}}.
\end{promptbox}

\begin{promptbox}[colback=darkyellow!20!white, colframe=darkyellow!80]{Qwen3-Omni}
You are a professional \textbf{\{src\_lang\}}-to-\textbf{\{tgt\_lang\}} translator. Your goal is to accurately convey the meaning and nuances of the original \textbf{\{src\_lang\}} speech while adhering to \textbf{\{tgt\_lang\}} grammar, vocabulary, and cultural sensitivities. Use precise terminology and a tone appropriate for academic or instructional materials. Produce only the \textbf{\{tgt\_lang\}} translation, without any additional explanations or commentary. Please translate the provided \textbf{\{src\_lang\}} speech into \textbf{\{tgt\_lang\}}:
\end{promptbox}

The inference is conducted on a mixed environment with NVIDIA A40 40GB, and NVIDIA L40S 48GB. A single GPU is used. On average, a single run takes $\sim$1-2 hours for SeamlessM4T, $\sim$4-5 hours for Phi4-Multimodal, $\sim$25-26 hours for Qwen3-Omni.

\section{Numerical Results}
\label{app:final-table}

Tables \ref{tab:punct-vs-fix} and \ref{tab:final-tab} show numerical results for Figures \ref{fig:punct-vs-fix} and \ref{fig:final}, together with quality and latency complementary metrics (BLEU score and StreamLAAL, respectively).


\section{AI Use Statement}
AI tools such as ChatGPT have been used for polishing the writing of the paper, and Codex has been used only for debugging purposes.

\begin{table*}[]
\centering
\resizebox{0.9\textwidth}{!}{%
\setlength{\tabcolsep}{10pt}
\begin{tabular}{lrrrrrr}
\toprule
\textbf{Method}                           &  $f$ & \multicolumn{1}{l}{\textbf{BLEU$\uparrow$}} & \multicolumn{1}{l}{\textbf{COMET$\uparrow$}} & \multicolumn{1}{l}{\textbf{LongYAAL$\downarrow$}} & \multicolumn{1}{l}{\textbf{LongLAAL$\downarrow$}} & \multicolumn{1}{l}{\textbf{empty\%$\downarrow$}} \\
\midrule
\multirow{5}{*}{punctuation}      & 5                          & 30.12                    & 0.7317                    & 1511                         & 1599                         & 0.43\%                                     \\
                                  & 10                         & 32.23                    & 0.7476                    & 1976                         & 2150                         & 0.21\%                                     \\
                                  & 15                         & 32.80                    & 0.7552                    & 2389                         & 2487                         & 0.43\%                                     \\
                                  & 20                         & 33.98                    & 0.7593                    & 2835                         & 2947                         & 0.43\%                                     \\
                                  & 25                         & 34.70                    & 0.7673                    & 3219                         & 3460                         & 0.21\%                                     \\
\midrule
\multirow{5}{*}{fixed words ($N=10$)} & 5                          & 29.02                    & 0.7193                    & 1476                         & 1560                         & 0.00\%                                     \\
                                  & 10                         & 31.29                    & 0.7362                    & 1957                         & 2050                         & 0.00\%                                     \\
                                  & 15                         & 32.95                    & 0.7525                    & 2374                         & 2471                         & 0.00\%                                     \\
                                  & 20                         & 33.74                    & 0.7537                    & 2790                         & 2891                         & 0.00\%                                     \\
                                  & 25                         & 33.97                    & 0.755                     & 3178                         & 3305                         & 0.43\%                                     \\
\multirow{5}{*}{fixed words ($N=20$)} & 5                          & 29.37                    & 0.7339                    & 1476                         & 1619                         & 0.21\%                                     \\
                                  & 10                         & 27.01                    & 0.7274                    & 1980                         & 3825                         & 0.21\%                                     \\
                                  & 15                         & 31.13                    & 0.7519                    & 2383                         & 2927                         & 0.00\%                                     \\
                                  & 20                         & 30.30                    & 0.7375                    & 2748                         & 3511                         & 0.21\%                                     \\
                                  & 25                         & 34.36                    & 0.7708                    & 3175                         & 3317                         & 0.43\%                                     \\
\multirow{5}{*}{fixed words ($N=30$)} & 5                          & 16.74                    & 0.7119                    & 1516                         & 7791                         & 0.21\%                                     \\
                                  & 10                         & 32.95                    & 0.7487                    & 1966                         & 2049                         & 0.43\%                                     \\
                                  & 15                         & 33.74                    & 0.759                     & 2379                         & 2476                         & 0.43\%                                     \\
                                  & 20                         & 34.08                    & 0.764                     & 2797                         & 2897                         & 0.43\%                                     \\
                                  & 25                         & 34.57                    & 0.7691                    & 3200                         & 3312                         & 0.21\%                              \\
\bottomrule
\end{tabular}
}
\caption{Numerical results for systems reported in Figure \ref{fig:punct-vs-fix} on ACL 60/60.}
    \label{tab:punct-vs-fix}
\end{table*}

\begin{table*}[!ht]
    \centering
\resizebox{0.9\textwidth}{!}{%
\setlength{\tabcolsep}{10pt}
\begin{tabular}{l|r|rrrrr}
\toprule
\textbf{Model}                           &  $f$ & \multicolumn{1}{l}{\textbf{BLEU$\uparrow$}} & \multicolumn{1}{l}{\textbf{COMET$\uparrow$}} & \multicolumn{1}{l}{\textbf{LongYAAL$\downarrow$}} & \multicolumn{1}{l}{\textbf{LongLAAL$\downarrow$}} & \multicolumn{1}{l}{\textbf{empty\%$\downarrow$}} \\
\midrule
\multicolumn{7}{c}{\textbf{\textit{en-de}}}       \\
\midrule
\multirow{3}{*}{SeamlessM4T}    & 4                          & 21.95                    & 0.6857                    & 1682                         & 1825                         & 1.09                                     \\
                                & 8                          & 23.77                    & 0.7026                    & 2265                         & 2379                         & 1.20                                     \\
                                & 12                         & 24.58                    & 0.7124                    & 3157                         & 3292                         & 0.98                                     \\
\hline
\multirow{3}{*}{Phi4Multimodal} & 5                          & 15.45                    & 0.7091                    & 1338                         & 7739                         & 0.44                                     \\
                                & 15                         & 29.40                    & 0.7602                    & 2240                         & 2334                         & 0.33                                     \\
                                & 25                         & 30.88                    & 0.7682                    & 3044                         & 3161                         & 0.22                                     \\
\hline
\multirow{3}{*}{Qwen3-Omni}     & 5                          & 24.59                    & 0.7392                    & 725                          & 896                          & 9.25                                     \\
                                & 15                         & 26.49                    & 0.7884                    & 2744                         & 3959                         & 3.70                                     \\
                                & 25                         & 28.18                    & 0.7911                   & 3749                         & 4889                         & 0.44                                     \\
\midrule
\multicolumn{7}{c}{\textbf{\textit{en-it}}}       \\
\midrule
\multirow{3}{*}{SeamlessM4T}    & 4                          & 33.37                    & 0.7592                    & 1631                         & 1735                         & 0.76                                     \\
                                & 8                          & 34.58                    & 0.7663                    & 2238                         & 2319                         & 0.65                                     \\
                                & 12                         & 36.48                    & 0.7725                    & 3064                         & 3209                         & 0.65                                     \\
\hline
\multirow{3}{*}{Phi4Multimodal} & 5                          & 29.73                    & 0.7707                    & 1327                         & 2950                         & 1.20                                     \\
                                & 15                         & 32.26                    & 0.787                     & 2174                         & 4051                         & 1.20                                     \\
                                & 25                         & 33.68                    & 0.7985                    & 3024                         & 4891                         & 0.11                                     \\
\hline
\multirow{3}{*}{Qwen3-Omni}     & 5                          & 34.78                    & 0.801                     & 72                           & 998                          & 3.81                                     \\
                                & 15                         & 37.26                    & 0.8086                    & 620                          & 1926                         & 3.70                                     \\
                                & 25                         & 38.14                    & 0.8282                    & 3806                         & 5466                         & 0.33                                    \\
\bottomrule
\end{tabular}
}
    \caption{Numerical results for systems reported in Figure \ref{fig:final} on MCIF.}
    \label{tab:final-tab}
\end{table*}

\end{document}